\documentclass[runningheads]{llncs}
\usepackage{graphicx}
\usepackage{booktabs}
\usepackage{amsmath}
\usepackage{amssymb}
\begin{document}
	\title{A Neural Approach to Ordinal Regression for the Preventive Assessment of Developmental Dyslexia}
	\titlerunning{A Neural Approach to Ordinal Regression for...}
	\author{Francisco J. Martinez-Murcia\inst{1} \and Andres Ortiz\inst{1} \and Marco A. Formoso\inst{1} \and 
		Miguel Lopez-Zamora\inst{2}\and
		Juan Luis Luque\inst{2}\and Almudena Gimenez\inst{2}}
	\authorrunning{F.J. Martinez-Murcia et al.}
	%
	\institute{Dept. of Communications Engineering, University of M\'alaga, Spain\\
		\email{fjmm@ic.uma.es} \and Dept. of Developmental and Educational Psychology, University of M\'alaga, Spain}
	\maketitle              
	\begin{abstract}
		Developmental Dyslexia (DD) is a learning disability related to the acquisition of reading skills that affects about 5\% of the population. DD can have an enormous impact on the intellectual and personal development of affected children, so early detection is key to implementing preventive strategies for teaching language. Research has shown that there may be biological underpinnings to DD that affect phoneme processing, and hence these symptoms may be identifiable before reading ability is acquired, allowing for early intervention. 
		In this paper we propose a new methodology to assess the risk of DD before students learn to read. For this purpose, we propose a mixed neural model that calculates risk levels of dyslexia from tests that can be completed at the age of 5 years. Our method first trains an auto-encoder, and then combines the trained encoder with an optimized ordinal regression neural network devised to ensure consistency of predictions. Our experiments show that the system is able to detect unaffected subjects two years before it can assess the risk of DD based mainly on phonological processing, giving a specificity of 0.969 and a correct rate of more than 0.92. In addition, the trained encoder can be used to transform test results into an interpretable subject spatial distribution that facilitates risk assessment and validates methodology.  
	\end{abstract}
	\section{Introduction}
	The Developmental Dyslexia (DD) is a learning disability that hinders the acquisition of reading skills, affecting roughly a 5\% of the population \cite{Peterson2012}. It is characterized by difficulties in reading, unreadable handwriting, letter migration and common misspelling, that can affect the intellectual and personal development of affected children \cite{Thompson2015}. A variety of learning methodologies exist that can positively impact the reading abilities of affected children. However, the diagnosis is generally associated to reading, and therefore it limits the minimum age, which may be of fundamental impact to apply preventive treatments. 
	
	In the last years, many works are pointing to common biological underpinnings that may cause this disability. Specifically, an incorrect phonological processing may lay behind some problems associated with DD, causing an abnormal encoding of words in memory \cite{kimppaImpairedNeuralMechanism2018,goswamiNeuralOscillationsPerspective2019,goswamiSpeechRhythmLanguage2019}. These symptoms may be identifiable before the subjects acquire the ability to read, allowing for a preventive intervention that favours the reading competence in those in risk for DD.
	
	Data decomposition is often used in machine learning both for dimensionality reduction and interpretation of the methodology. Many methodologies exist, among then the very popular Principal Component Analysis (PCA) \cite{Spetsieris} or Independent Component Analysis (ICA) \cite{Bartlett2002,martinez-murciaFunctionalActivityMaps2013}. In contrast to these techniques, whose model of the data is a linear combination of hidden variables, there exist manifold learning techniques that allow to discover more complex combinations. Some of them have been widely used in the literature, such as Isomap, the t-distributed stochastic neighbour embedding (t-SNE) or more recently, autoencoders \cite{chen2017deep,8737996}. Autoencoders are a powerful and versatile tool used in many works to yield a data-driven distribution of the data, allowing for correlation with continuous variables and classification, e.g. in Alzheimer's Disease \cite{8737996}. 
	
	Typically, machine learning is often thought of as a variety of methods for classification and regression. Moreover, most methodologies that deal with any kind of diagnosis tend to make use of classifiers, whereas those dealing with continuous assessment --e.g., cognitive tests-- make use of regression. However, there is an intermediate problem, for which there are not so many alternatives available: risk assessment \cite{li2007ordinal}. In the case of DD, there exist an arbitrary scale ranging from 0 to 4, in which no intermediate values are available. The risk scale is not continuous in nature, but in contrast to multi-class classification the outcomes still depend on each other: level 3 implies a higher risk than 2, but smaller than 4. There is an ordinal nature in these problems, and that is why we use ordinal regression to tackle the problem. 
	
	Many ordinal regression methods have been proposed. The most widespread consists on dividing the grading problem in a series of binary classifiers, each of which indicates if a certain threshold has been surpassed \cite{li2007ordinal,niu2016ordinal}. However, most of these systems deal with inconsistency in the classifier when the training complexity increases. That is, some binary classifiers may indicate the grade is above a given threshold, whereas others may not. In \cite{caoRankconsistentOrdinalRegression2019}, the authors propose a Consistent Rank Logits (CORAL) ordinal regression to implement the binary classifiers with parameter sharing in the weights of the last layers, but with individual biases in each neuron, accomplishing theoretical classifier consistence. 
	
	In this paper, we present a novel methodology to predict the risk of DD in 5 year old individuals based on the outcomes of tests designed by expert psychologists. These subjects were followed over 4 years (from 5 to 8 years old), until a consistent DD risk evaluation was performed at age 7. We apply autoencoders for obtaining a feature modelling of the test outcomes and then a ordinal neural regressor that tries to predict the risk levels using the data at age 7. The dataset and the complete methodology is introduced at Section~\ref{sec:methods}, the results are presented at Sec.~\ref{sec:results} and discussed at Sec.~\ref{sec:discussion}. Finally, conclusions about this work are drawn at Section~\ref{sec:conclusions}.

	\section{Material and Methods}\label{sec:methods}
	\subsection{The LEEDUCA Study}
	The LEEDUCA project is a study for the assessment of specific learning difficulties of reading -dyslexia- and their evolution during infancy \cite{martinez2019periodogram}. It implements a Response to Intervention (RtI) system that has been applied for 20 years in the US and 10 years in Finland. The system applies a dynamic evaluation three times a year from 4 to 8 years to large population samples. Specifically, the control and experimental groups came from a cohort from different schools in the south of Spain, following evaluation from 5 to 8 years, via the standard criteria used in similar studies and the Special Education School Services (SESS). 
	
	\subsection{Data and Preprocessing}
	The data from the LEEDUCA study comprises a battery of tests spanning from 5 to 8 years old children at school. These tests are adapted to the age and educational level of the students (e.g., when they cannot read at 5 years), and therefore it is difficult to establish any longitudinal processing. As stated, we only use the 33 tests performed at 5 years --when students cannot read-- to predict the risk of DD at 7 years. The risk grades were set in function of the number and grade of the scores in assessments of four major categories: Phonological Route, Visual Route, Text Fluidity and Text Comprehension. Anomalous values were set based on percentile ($p$) values, in which $p<30$, $p<20$ and $p<30$ were set to grade 1, 2 and 3. Afterwards, these abnormality grades were averaged over all assessments and categories, and the final risk was estimated according to this average value, in a scale of 0 to 4, depending on the intervals where it laid $(-\infty, 0), (0,0.5), (0.5, 1.5), (1.5, 2.5), (2.5, \infty)$. In the end, the number of subjects in each grade is 5, 331, 270, 50 and 10 for grades 0, 1, 2, 3 and 4. 
	
	All subjects lacking more than 5 test results in the tests were deleted from the set. That left us with 572 subjects with evaluations at 5 and 7. K-Nearest-Neighbour (KNN) imputation \cite{troyanskayaMissingValueEstimation2001} with the two closest neighbours was used to generate valid values for those missing less than 5 results. Finally, the data was scaled to the range $[0,1]$, estimating the minimum and maximum values from the training subset.
	
	\subsection{Denoising Autoencoder}\label{sec:autoencoder}
	Autoencoders (AEs) are a specific type of neural encoder-decoder architecture. It consists of a feed-forward neural network that reduces dimensionality (encoder), directly connected to a inverse network (a decoder, usually symmetric with the encoder) that increases dimensionality to reconstruct the original shape. Then, the network is trained to minimize the error between the input and the output. A typical variation is the denoising AE (DAE), in which the input is corrupted with noise and the network is expected to provide the original input, without noise, which is sometimes considered a regularization procedure. No further regularization was used. 
	
	In this work, we propose a hybrid model that trains the autoencoder and then reuses the encoder part to perform dimensionality reduction, as in \cite{8737996}. The precise architecture uses symmetric encoder and decoder modules. There are three layers of $N$, 64 and 3 neurons for the encoder and 3, 64 and $N$ for the decoder (where $N$ is the number of tests included). We used 3 neurons in the $Z$-layer to favour a visual interpretation of the results in a three-dimensional space, and 64 neurons in the intermediate layers of the encoder and decoder were chosen after a careful systematic test of accuracy and visualization, in powers of 2. A higher number of neurons led to overfitting and lower explainability of the representation, and a smaller number of neurons yielded lower performance.
	Batch normalization is used for speeding up the convergence and the activation function for layers 1, 2, 4 and 5 is ELU. The intermediate layer (usually known as Z-layer) and output layers have linear activation. For training we use the Mean Squared Error (MSE) between the input and output data as loss, and the Adamax optimizer \cite{kingma2014adam}. 
	
	\subsection{Ordinal Neural Regression}\label{sec:coral}
	To perform ordinal regression, we use the Consistent Rank Logits (CORAL) approach proposed in \cite{caoRankconsistentOrdinalRegression2019}. CORAL is devised to create an ordinal regression framework with theoretical guarantees for classifier consistency, in contrast to other methods in the literature \cite{niu2016ordinal}. The procedure consists of two major contributions. First, a label extension, by which the rank level $y_i$ is extended into $K-1$ binary labels $\lbrace t_{i,0}, \dots t_{i,K-1}\rbrace$ such that $t_{i,j}\in \lbrace0,1\rbrace$ indicates whether $y_i$ exceeds a given rank ($y_i > r_k$), as in \cite{niu2016ordinal}. This is implemented at the output layer of the regression network, via a layer with $K-1$ binary neuron classifiers sharing the same weight parameter but independent bias units, which according to \cite{caoRankconsistentOrdinalRegression2019} solves the inconsistency problem among predicted binary responses. The predicted rank is obtained as:
	\begin{equation}
	r_i = \sum_{j=0}^{K-2} o_{i,j}
	\end{equation}
	where $o_{i,j}$ is the output (linear activation) of the $j^{th}$ neuron for the $i^{th}$ subject, also known as logit. 
	
	The second key aspect of the CORAL regression is the loss function. To calculate the loss between $o_{i,j}$ and the target level $t_{i,j}$, the authors propose: 
	\begin{equation}
	\mathcal{L}(\mathbf{o}, \mathbf{l}) = \sum_n \sum_j t_{n,j} \log \left[s(o_{n,j})\right]+ (1-t_{n,j})(\log \left[s(o_{n,j})\right] - o_{n,j})
	\end{equation}
	where $s(\cdot)$ is the sigmoid function. An optional feature importance variable could multiply the second term to adjust for label prevalence, although adding it did not increased the performance significantly. Furthermore, since it also implied making assumptions about the real distribution of subjects, we chose not to use this importance term. 
	
	\subsection{Full Model: Architecture and Training}\label{sec:model}
\begin{figure}[htp]
		\centering
		\includegraphics[width=\textwidth]{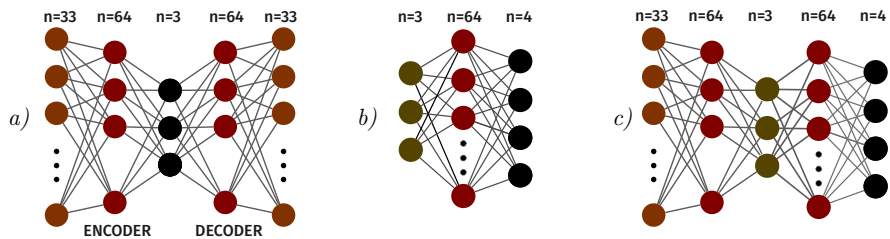}
		\caption{Schema of the a) Autoencoder for pre and re-training, b) Feed-forward network for CORAL ordinal regression, c) complete model.}
		\label{fig:model_properties}
	\end{figure}

	The resulting model is a combination of the encoder part of a DAE and an ordinal neural regressor, a 3-layer feed-forward network that uses the CORAL framework. The model architecture is displayed in more detail at Figure~\ref{fig:model_properties}.a), b) and c). 
	
	The cost functions for the encoder and the regressor are defined at Sections~\ref{sec:autoencoder} and \ref{sec:coral}, and Adam with $lr=0.01$ is used to train the whole system (with or without locking the encoder). We applied early stopping in both cases, that is, the training was stopped after 150 epochs if there was no improvement in validation loss, in order to retain the best model. 
	
	\subsection{Evaluation}\label{sec:evaluation}
	The models have been tested under a 10-fold stratified cross-validation (CV) scenario \cite{Kohavi1995a}. A large variety of performance measures were obtained, specifically: correct rate, per-class correct rate (for multi-class classification), sensitivity, specificity, precision and F1-score, and their corresponding standard deviation (STD) over all cross-validation folds. Finally, the balanced accuracy is defined as the average correct rate over all classes \cite{8737996}, which is the most representative measure in such an imbalanced problem.
	
	For evaluating the relative contribution of each variable to the final outcome (either the manifold distribution, risk estimation or classification), we use a randomization procedure based on Permutation Importance (PI) \cite{oldenIlluminatingBlackBox2002}. We iteratively set each variable to zero, and re-tested the trained regressor obtaining a performance estimation in each CV fold. Afterwards, we compare those performances to the regressor using all variables. In this framework, a larger performance loss implies a larger influence of that variable in the model.  
	
	Finally, we define the following models to be compared in our work: 
	\begin{itemize}
		\item \textbf{PCA}. A model composed of a decomposition of the dataset using Principal Component Analysis and a CORAL regression (Fig.~\ref{fig:model_properties}.b) on the component scores for each subject. Note that only the training subset is used to create the PCA model and project the test set. 
		\item \textbf{Pretraining}. The proposed model in which the autoencoder (Fig.~\ref{fig:model_properties}.a) is first trained, and then the model is built with the neural regressor (Fig.~\ref{fig:model_properties}.b) and the pre-trained encoder (Fig.~\ref{fig:model_properties}.c). The encoder is locked and only the neural regressor is trained.
		\item \textbf{Retraining}. The AE is pre-trained as in the previous model (Fig. \ref{fig:model_properties}.a), but this time, the full model (Fig.~\ref{fig:model_properties}.c) including the encoder and the neural regressor are trained simultaneously. 
	\end{itemize}
	
	\section{Results}\label{sec:results}
	After training and testing the models defined in Section \ref{sec:evaluation}, we measured first the multi-level performance; that means, we provide the accuracy for the different DD risk levels, the overall correct rate and the balanced accuracy. These results are presented at Table~\ref{tab:multi-level}.
	
	\begin{table}[htp]
		\centering
		\caption{Results for the ordinal regression, including accuracy per level, balanced accuracy and overall correct rate for the PCA, pre- and re-training models.}
		\label{tab:multi-level}
		\begin{tabular}{lrrr}
			\toprule
			{}                 &                   PCA &           pre-training &           re-training \\ \midrule
			Level 0 (acc.)     &          0.000 [0.00] & \textbf{ 0.143 [0.14]} &          0.067 [0.14] \\
			Level 1 (acc.)     & \textbf{0.822 [0.13]} &           0.633 [0.17] &          0.615 [0.15] \\
			Level 2 (acc.)     & \textbf{0.433 [0.18]} &           0.376 [0.09] &          0.395 [0.13] \\
			Level 3 (acc.)     &          0.058 [0.12] &           0.225 [0.26] & \textbf{0.442 [0.14]} \\
			Level 4 (acc.)     &          0.000 [0.00] &           0.000 [0.00] &          0.000 [0.00] \\
			Correct Rate [STD] & \textbf{0.575 [0.06]} &           0.481 [0.07] &          0.484 [0.08] \\
			Balanced Acc.      &          0.309 [0.05] &           0.321 [0.05] & \textbf{0.357 [0.06]} \\ \bottomrule
		\end{tabular}
	\end{table}

\begin{figure}[htp]
		\centering
		\includegraphics[width=\textwidth]{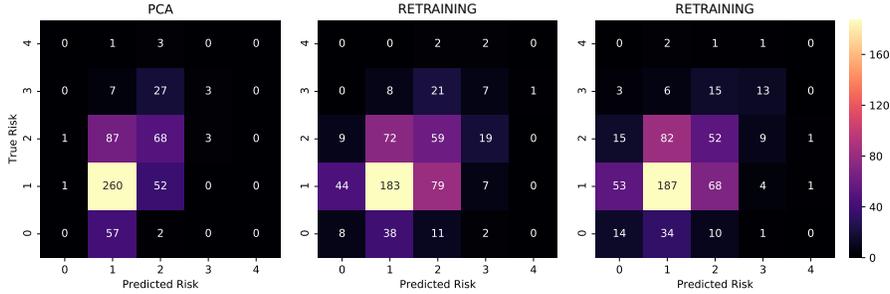}
		\caption{Confusion matrices for the three models evaluated.}
		\label{fig:compare_confmat}
	\end{figure}

	Regarding the per-level accuracy, we observe that PCA is good in general for obtaining a fair overall correct rate (0.575) when compared to the pre-training model (0.481) and the re-trained model(0.484). However, the PCA fails to account for the less-prevalent levels (3 and 4, the ones associated with high risk of DD), which is precisely the main objective of this paper. When looking at this, the pre-training model at least detects a small proportion (0.225) of level 3 subjects and also level 0, but failing to account for level 4, whereas the re-trained model detects a larger amount (0.442) of these levels, at the cost of mistaking some level 1 and 2 subjects. This is reflected on the balanced accuracy, which is higher in the case of the re-training model, but also can be seen in more detail at the confusion matrices, displayed at Figure~\ref{fig:compare_confmat}. 
	
	There we see how the re-training model is the best in grading extreme-level subjects (levels 0, 3 and 4) correctly. This is far more evident in the case of the levels more associated to dyslexia (3 and 4), which are far better in the DAE + ordinal regression models than in the PCA.
	
	However, there is another approach to the problem: the one that considers only the detection of subjects in high risk of DD (a risk level $\ge3$). Derived from the same model, the results of this binary scenario are presented at Table~\ref{tab:results_binary}. 
	
	\begin{table}[htp]
		\caption{Performance results for binary classification of the highest levels ($r_i\ge3$)}
		\label{tab:results_binary}
		\centering
		\begin{tabular}{lrrr}
			\toprule
			{} &         PCA &         pre-training &         re-training \\
			\midrule
				Correct Rate &  \textbf{0.934 [0.012]} & 0.898 [0.052] &  0.927 [0.041] \\
				Sensitivity &    0.050 [0.110] &  0.225 [0.281] &   \textbf{0.442 [0.153]} \\
				Specificity & \textbf{0.998 [0.006]} & 0.946 [0.059] &  0.962 [0.037] \\
				Precision   &   0.667 [0.580] &  0.446 [0.482] &  \textbf{0.909 [0.091]} \\
				F1-Score      &      0.400 [0.000] &  0.572 [0.193] &  \textbf{ 0.585 [0.150]} \\
				Balanced Acc.      &  0.524 [0.053] &  0.586 [0.131] &  \textbf{0.702 [0.086]} \\
			\bottomrule 
		\end{tabular}
	\end{table}
	
	The overall correct rate in this case is astounding, over 90\% in all cases. However, this is again due to the lower prevalence of DD. When digging deeper into the measures, we observe that the sensitivity and specificity, as well as the balanced accuracy, offer a clearer sight of the performance of each model. Particularly, the larger specificity is achieved again by the PCA model (it is the best in discarding subjects with no risk of DD). However, the sensitivity of the model is 0.050, close to labelling all subjects as low risk. The DAE+Regressor models offer larger sensitivity (over 0.2), again with the re-training model the one achieving better results, with a sensitivity of 0.442, and a precision of 0.909, and a total balanced accuracy of 0.702, which is fairly good for this application. Furthermore, the specificity of this method (0.927) is excellent, indicating that the system hardly misses any subject in risk of DD. Given that our purpose is to perform a preventive intervention on subjects in risk, a good trade-off between high specificity and the best possible sensitivity is a sensible approach.
	
	\section{Discussion}\label{sec:discussion}
	The results of evaluating the model show that it is possible to predict DD when students are 5 years old, before they have learnt to read. This is a fundamental advance in preventive treatment, allowing for an earlier detection of this learning disability and making it possible to apply a prevention program. 
	
	Since the LEEDUCA study has a large cohort that has been repeatedly evaluated over the years, many subjects are available for our study. Thus, neural network architectures may be applied to the problem. The combination of a representation modelling approach (the DAE) that has been applied to other data analysis pipelines \cite{8737996}, and the flexibility of neural regression seems to be informative enough to automatically grade the risk of DD and provide a set of subjects to which the preventive program could be applied. Moreover, this methodology gives us larger insight into which tests are more predictive of future reading disability, at the same time that it provides a deeper insight into the data, revealing a self-supervised data decomposition via the autoencoder that allows for a bi- and tri-dimensional representation of the dataset. 
	
\begin{figure}[htp]
		\centering
		\includegraphics[width=\textwidth]{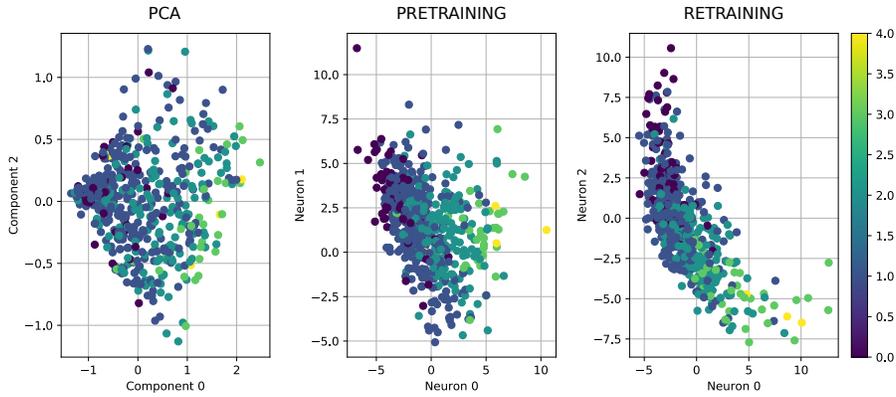}
		\caption{Distribution of the points at the output of the PCA (for the baseline system) and the AE (for the pre- and re-training systems). Note the self-supervised distribution of the gradings in the later models.}
		\label{fig:points_manifold}
	\end{figure}

	When exploring this projection onto a two-dimensional space, and comparing the three models, we obtain Figure~\ref{fig:points_manifold}. There we can see that all three methods project the subjects (each point) in a spatial coordinate related to the DD risk. However, these distributions differ. In the case of PCA, the risk increases with component 0 (the first PCA component), but individuals are sparsely located, and the levels are very mixed, which was expected for a linear approach. For the DAE + regressor models, the levels are more distinguishable. In the case of the pre-training model, it resembles the PCA model with an increasing risk over neuron 0, but this time, the levels are clustered together. However, the subjects at higher level are still very mixed, making it difficult for the regressor to correctly assign risk levels. The re-training model, however, is forming a manifold that resembles a curve starting at subjects with risk 0-1 and relatively increasing up to the furthest subjects, those with risk level 3 and 4. This proves that there exist a relationship between the test outcomes and the risk of developing DD that is better modelled by the re-training model, generating more accurate predictions of risk. 
	
\begin{figure}[htp]
		\centering
		\includegraphics[width=0.7\textwidth]{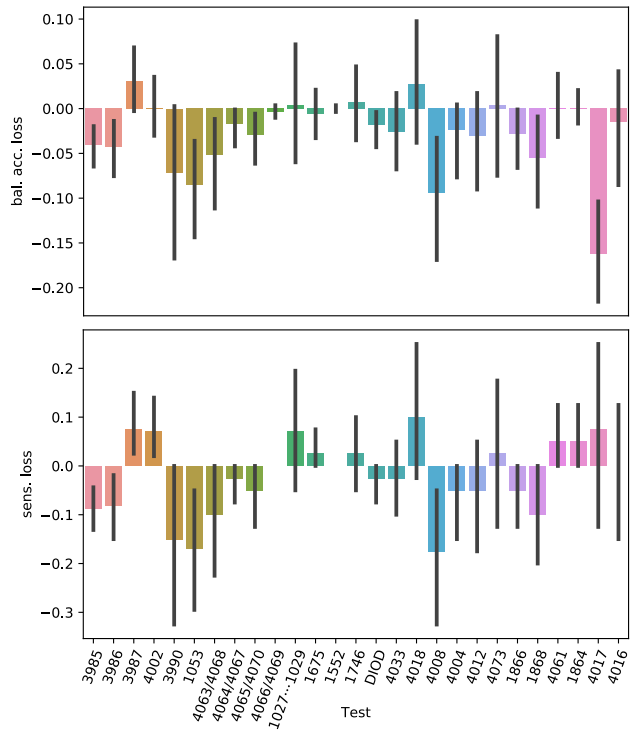}
		\caption{Relative importance of the different variables in the re-training system, computed using the sensitivity and balanced accuracy loss.}
		\label{fig:perm_importance}
	\end{figure}

	Focusing on this model, it may be very useful to assess which input variables cause larger changes in the overall balanced accuracy and sensitivity. To do so, we use the PI algorithm (see Sec.~\ref{sec:evaluation}) that helps us to visualize the relative influence of each variable. This importance is shown at Figure~\ref{fig:perm_importance}. 
	
	In Fig.~~\ref{fig:perm_importance} the behaviour is consistent in most variables, regardless of how the loss is measured (balanced accuracy or sensitivity in the binary classification, DD detection), except for two relevant cases, test variables 4017 and 4002. Those correspond respectively to a test to label Arabic numerals and a test for verbal memory according to the phonological hypothesis. For the remaining variables, the larger performance loss is for variables 3990, 1053, 4063/4068 and 4008. The last one corresponds to an evaluation of the `sustained attention', within the category of executive functions. All the remaining correspond to the phonological hypothesis, specifically for trials that evaluate the Lexical access speed in colours (3990), numbers (1053) and objects in syllables (4063/4068). This is consistent with the current scientific understanding of DD in the literature. In fact, an incorrect phonological processing is one of the most accepted causes of DD in the scientific community \cite{shaywitz2008education}, causing an abnormal encoding of words in memory. 
	
	In summary, we developed a novel methodology intended to accurately grade subjects two years before the first DD risk evaluations. It uses a self-supervised decomposition via denoising autoencoders plus a neural ordinal regression following the CORAL methodology. The results prove that the methodology is useful for an early screening, achieving high values of specificity that could lead to non-invasive preventive methodologies that allow a more efficient development of reading skills. 
	
	\section{Conclusions}\label{sec:conclusions}
	In this work, we have developed a neural system that combines a denoising autoencoder with the theoretical guarantees of the Consistent Rank Logits (CORAL) neural regression, allowing for a model that is able to predict the risk of Developmental Dyslexia two years before first assessing the reading abilities. The system combines a pre-training of the autoencoder, and then connecting the output of the encoder to a neural perceptron that uses parameter sharing at the output as in the CORAL ordinal regression framework, yielding a specificity of 0.969 and correct rate of over 0.92. The system outputs risk level values similar to the ones assessed at age 7 using just the test outcomes at age 5, based mainly on phonological processing. The system proved its ability in detecting non-affected and yielding a subset of candidates for preventive --non invasive-- language teaching modalities, allowing a visual interpretation by transforming the battery of tests into a manifold related to the risk levels of dyslexia, validating the methodology. 
	
	\section*{Acknowledgements}
	This work was partly supported by the MINECO/FEDER under RTI2018-098913-B-I00, PSI2015-65848-R and 
	PGC2018-098813-B-C32 projects. Work by F.J.M.M. was supported by the MICINN ``Juan de la Cierva - Formaci\'on'' FJCI-2017-33022.
	%
	%
	%
	\bibliographystyle{splncs04}
	\bibliography{bibliography.bib}
\end{document}